\documentclass[11pt]{article}

\usepackage{arxiv}

\usepackage{times}
\usepackage{latexsym}
\usepackage[numbers]{natbib}

\usepackage[utf8]{inputenc}
\usepackage{url}
\usepackage{hyperref}
\hypersetup{colorlinks,allcolors=black} 

\usepackage{graphicx}
\DeclareGraphicsExtensions{.png,.pdf}
\usepackage{subcaption}
\usepackage{multirow}
\usepackage{booktabs}

\usepackage{float}
\usepackage{longtable}

\usepackage{amsmath}

\title{From Encoding to Generation: Towards Empathetic Large Language Models}
\author{Samhita Bollepally \\
  \texttt{samhita.bollepally@tamu.edu} \\
  \And
   Takashi Yamauchi \\
  \texttt{takashi-yamauchi@tamu.edu} \\
  \And
  Marissa West\\
  \texttt{mroowest03@tamu.edu}\\
}

\begin{document}
\maketitle
\begin{abstract}
Empathetic interaction involves two distinguishable capacities: representing another person's emotional state (cognitive empathy) and generating an appropriate response to that state (affective empathy). We apply this distinction to examine how emotion-related information is internally organized, selectively attended to, and carried through response generation in Llama-3.2-3B-Instruct across 10 emotion categories from the EmpatheticDialogues dataset. Layer-wise linear probing revealed a broad middle-layer region of peak emotion-category decodability, with the layer-11 L2 probe achieving 63.6\% held-out accuracy against a 10\% chance level and a 54.6\% TF-IDF lexical baseline. During response generation, emotion-related user tokens received 1.89 times as much attention per token as other utterance tokens, concentrated across layers 10--15. Target-emotion alignment then decreased from a mean cosine similarity of 0.239 for input representations to 0.125 for generated responses, with positive drift in 77\% of examples and the largest shifts for \emph{angry} and \emph{furious}. Emotion-category representation, attention allocation, and response-level alignment therefore emerge as distinguishable components of model processing that can be analyzed through a functional cognitive–affective empathy framework. These findings provide a foundation for monitoring emotion representations during generation and leveraging models’ existing affective structure to support more naturally empathetic responses without relying on external prompting or explicit emotion conditioning.
\end{abstract}

\section{Introduction}

Empathy is a foundational capacity in human social interaction, encompassing two related but dissociable components. \emph{Cognitive empathy} is the capacity to identify and encode what another person is feeling. \emph{Affective empathy} is the capacity to generate a response that is appropriately aligned with or responsive to that perceived state. These components dissociate in humans; for example, a person may recognize others' distress without producing a supportive response or may respond emotionally without fully understanding the other's situation. Because empathy unfolds across both representation and response, studying it requires examining how emotional information is internally organized and how that organization shapes behavior.

Large language models (LLMs) are now routinely deployed in emotionally sensitive contexts like mental health support \citep{fitzpatrick2017}, peer counseling \citep{sharma2021}, and open-domain conversation \citep{rashkin2019towards}. Despite producing outputs rated as highly empathetic by human evaluators \citep{lee2024large}, it is not known whether this behavior is grounded in structured internal representations of user emotions or arises from the surface statistics of empathetic language. The cognitive-affective empathy distinction provides a concrete framework for this investigation: does the model represent the user's emotional state, preferentially attend to emotionally relevant input, and carry that representation into its response?

\subsection{Related Work}

Human empathy research distinguishes between representing others' emotional states and generating an appropriate response \citep{riess2017}. Both components rely on distributed neural patterns rather than isolated emotion centers \citep{fan2011, saarimaki2018distributed}, and structured emotion representations can be decoded from neural activity using multivariate pattern analysis \citep{baucom2012decoding}. The same decoding logic extends to transformer models: linear probes reveal what information is encoded at each layer, with different types of information emerging at different depths \citep{alain2017understanding, belinkov2022probing}.

LLMs trained with reinforcement learning from human feedback (RLHF) show a systematic positive-polarity bias, reliably generating positive emotional content while underserving negative emotions \citep{maheswaran2025}. This suggests that empathetic-seeming outputs can arise from alignment-induced response patterns rather than from robust internal representations of user emotional states. Prior work has examined emotion information in LLM representations, but less is known about how such representations relate to attention allocation and the representational profile of generated responses.

Motivated by the cognitive-affective empathy distinction, we investigate Llama-3.2-3B-Instruct using the EmpatheticDialogues benchmark \citep{rashkin2019towards} to address three questions:

\begin{enumerate}
    \item \textbf{Representation:} How is emotion-category information internally organized across the model's representational levels?
    \item \textbf{Attention:} Does emotionally relevant input receive preferential attention during response generation?
    \item \textbf{Response alignment:} How does emotion-related alignment change between the input representation and generated response, and how does this shift vary across emotion categories?
\end{enumerate}

To address these questions, three analyses are combined: (1) layer-wise L1- and L2-regularized linear probing of all 29 representational levels to assess emotion-category encoding; (2) layer-wise attention analysis across token-source categories to assess preferential processing of emotional input; and (3) paired comparison of input and generated-response representations against emotion-category contrastive directions derived at the probe-selected layer. Together, these analyses reveal that emotion information is encoded in a broad middle-layer region (63.6\% accuracy at layer 11), that emotionally relevant tokens receive preferential attention during generation (1.89 times enrichment), and that target-emotion alignment is only partially preserved in generated responses, with the largest decline for negative-valence emotions.

\section{Methodology}

This study examined how emotional information is represented and propagated in Llama-3.2-3B-Instruct \citep{grattafiori2024llama, meta2024llama32}, which comprises 29 representational levels: one initial token-embedding layer and 28 transformer decoder layers. The study combined three analyses: layer-wise probing of hidden-state representations, attention allocation during response generation, and input--response emotion-trajectory analysis.

\begin{figure}[!htbp]
\centering
\includegraphics[
  width=\columnwidth,
  height=0.25\textheight,
  keepaspectratio]{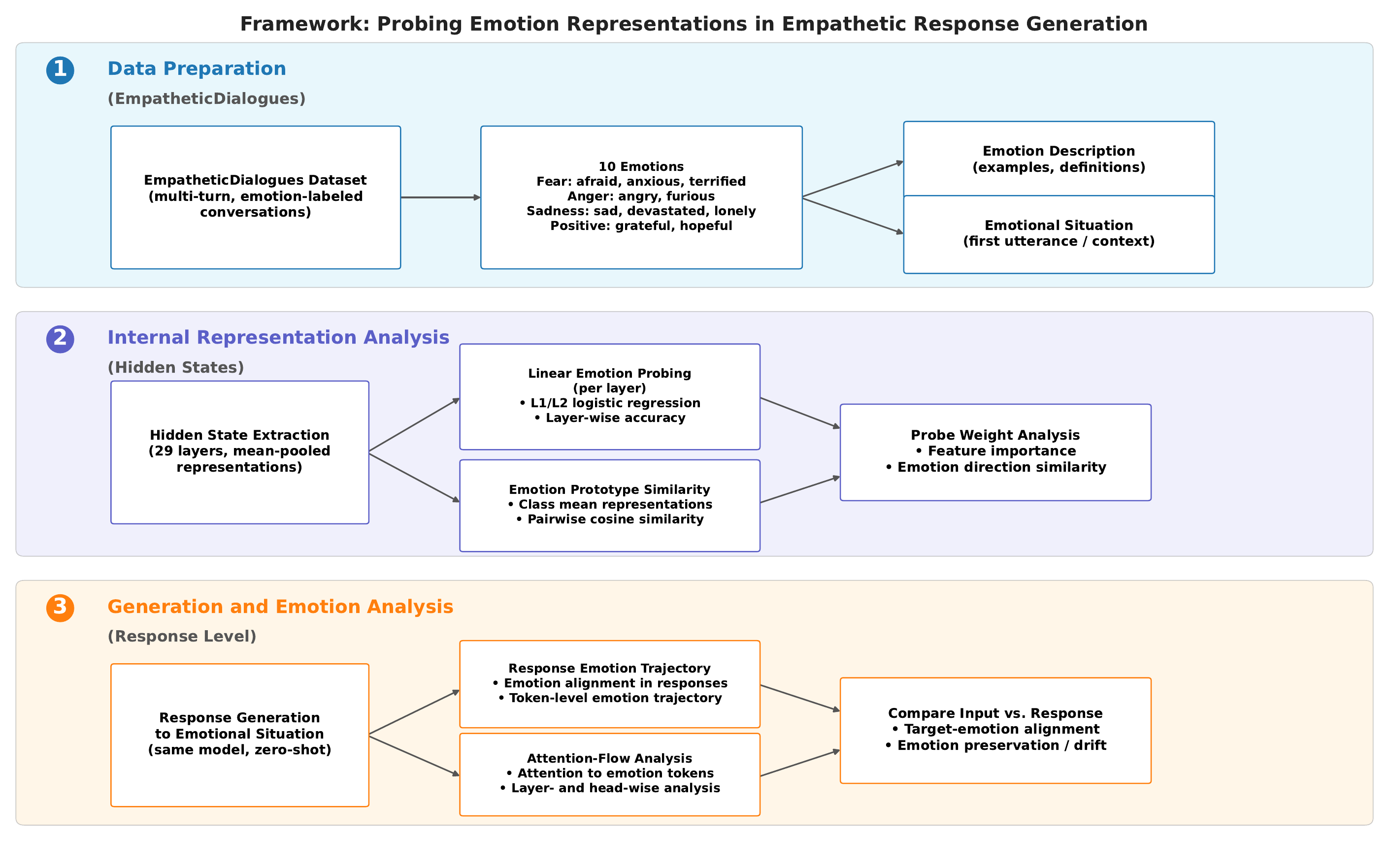}
\caption{Overview of the analysis pipeline. Hidden states are extracted from emotion-labeled situations (Set 1) to probe and construct prototypes. First utterances (Set 2) are fed to the model in a zero-shot response-generation setting; attention and hidden-state trajectories are recorded during generation.}
\label{fig:framework}
\end{figure}

\subsection{Materials}

The EmpatheticDialogues dataset \citep{rashkin2019towards}, a crowdsourced collection of approximately 25,000 open-domain multi-turn conversations grounded in 32 emotion categories, was used. Ten categories were selected: \emph{afraid}, \emph{angry}, \emph{anxious}, \emph{devastated}, \emph{lonely}, \emph{sad}, \emph{terrified}, \emph{furious}, \emph{grateful}, and \emph{hopeful}. These span fear- and anxiety-related states (afraid, anxious, terrified), anger-related states (angry, furious), sadness and isolation (sad, devastated, lonely), and positive states (grateful, hopeful), enabling analysis of both fine-grained distinctions and broader affective contrasts.

For each of the 10 emotions, 400 non-overlapping situation descriptions were sampled to form Set 1, and a separate 200 first utterances were sampled to form Set 2. Set 1 was used for representation-level probing and prototype construction; Set 2 was used for response generation and generation-level analyses. Representative examples from each set are shown in Tables~\ref{tab:set1-prompt-examples} and~\ref{tab:set2-utterance-examples}.

\begin{table*}[t]
\centering
\small
\begin{tabular}{p{0.12\textwidth} p{0.78\textwidth}}
\toprule
Emotion & Emotion description \\
\midrule

Afraid &
I was scared when I saw a mountain lion at night right in front of me. I ran to my tent. \\

\addlinespace

Angry &
Someone broke my window with a soccer ball. \\

\addlinespace

Anxious &
I drank all the coffee this morning and didn't share the whole pot with anyone. Now I feel very strange. \\

\addlinespace

Devastated &
Passing of my mother. \\

\addlinespace

Lonely &
My boyfriend went out of town for a few days, leaving me all alone at the apartment. I didn't have many friends here. \\

\addlinespace

Sad &
I just got laid off from my job. I need the paycheck, so this is not a good situation. \\

\addlinespace

Terrified &
A tornado came through my area last week \\

\addlinespace

Furious &
I was pissed when I saw someone put a dent in my car door. \\

\addlinespace

Grateful &
My mom let me move back in with her when my relationship didn't work out. I'm so thankful to have such a good mother. \\

\addlinespace

Hopeful &
I am going to the casino next weekend. I am entering a poker tournament and think I have a good chance. \\

\bottomrule
\end{tabular}
\caption{Examples of emotion-labeled descriptions from Set 1. Set 1 was used for hidden-state probing, prototype construction, and representational analyses. Text is reproduced from the dataset, with capitalization added only where necessary for consistent presentation.}
\label{tab:set1-prompt-examples}
\end{table*}

\subsection{Procedure}
\subsubsection{Hidden-State Extraction}

Each text sample from Set 1 was passed through the model separately. Hidden-state representations were extracted from all 29 levels and averaged using masked mean pooling to produce a 3,072-dimensional sequence-level vector, excluding padding and the BOS token \citep{reimers2019sentence, palma2025llamas}. For each emotion and layer, an emotion prototype was computed by averaging the 400 pooled vectors belonging to that emotion. A grand-mean prototype was computed across all Set 1 examples; the target-emotion contrastive direction was defined as the difference between the emotion prototype and the grand mean.

\subsubsection{Response Generation}

From the first 200 utterances available for each emotion in Set 2, 20 were sampled for response generation, yielding 200 generated responses in total and introduced into the model's chat template using a zero-shot system prompt, positioning the model as a medical assistant instructed to produce a one- to two-sentence response (full prompt in Appendix~\ref{app:generation-prompt}). Responses were generated autoregressively with nucleus sampling (temperature 0.7, top-$p$ 0.9, maximum 50 new tokens). Hidden states and attention distributions were recorded at every generation step.

\subsection{Analysis}
\subsubsection{Linear Emotion Probing}

To characterize how emotion-category information is organized across representational levels, Separate layer-wise multiclass logistic regression probes were trained. Each probe is trained independently at one of the 29 representational levels, following standard layer-wise probing methodology \citep{alain2017understanding, belinkov2022probing}. High probe accuracy at a given layer indicates that emotion-category information is linearly accessible from the representations at that depth, revealing where in the network emotion categories emerge.

Two regularization schemes were used to assess whether this information is sparse or distributed across hidden dimensions. \textbf{L1 (Lasso) regularization} produces sparse classifiers by driving many coefficients to exactly zero, testing whether accurate classification can be achieved using only a small subset of dimensions. \textbf{L2 (Ridge) regularization} penalizes large weights while retaining nonzero coefficients across many dimensions, allowing information distributed across the full hidden space to contribute (see Appendix~\ref{app:probing}). Comparing the two schemes, therefore, reveals whether emotion categories are encoded in a few localized features or spread across many partially redundant dimensions.

Set 1 is split once into stratified training (80\%) and test (20\%) partitions with a fixed random seed. For each layer and regularization scheme, the penalty strength is selected by stratified 5-fold cross-validation on the training partition. The best layer is selected separately for L1 and L2 based on cross-validated accuracy, and each selected probe is evaluated once on the held-out test partition. We report held-out accuracy and macro-F1 as primary metrics. This procedure ensures that probe performance reflects generalizable accessibility of emotion information rather than overfitting to the training data.

\subsubsection{Token-Group Attention Analysis}

To determine whether emotionally relevant input receives preferential processing during generation, we analyze how attention is allocated across token-source categories at each generation step. Attention distributions are aggregated over four mutually exclusive source groups: the BOS token, template tokens (system instructions and chat formatting), user-utterance tokens, and previously generated tokens. Emotion-related tokens are analyzed as a subset of the user-utterance group.

Emotion-related utterance tokens are identified using the contrastive directions derived from the Set 1 prototypes at layer 11, the layer at which linear emotion decoding peaked. A token is classified as emotion-related if the cosine similarity between its layer-11 hidden state and the target emotion's contrastive direction is at least 0.10. This links the attention analysis directly to the same representational space used in the probing analysis, so that emotion-related is defined by the model's own internal geometry rather than by surface lexical cues.

At each generation step, attention weights were summed within each source group. Because the groups differed in token count, mean attention per token was additionally computed for emotion-related and non-emotional utterance tokens. The per-token enrichment ratio was defined as the mean attention per emotion-related token divided by the mean attention per non-emotional utterance token. Its variation across layers was then examined to determine whether preferential attention occurred in the same middle-layer region in which emotion-category information was most decodable. This approach treats attention as an index of processing allocation, not as a causal explanation of output \citep{jain2019attention, wiegreffe2019attention}. It allows us to test whether the model actively prioritizes emotional content during response construction.

\subsubsection{Response Emotion Trajectory}

To measure how emotional information from the input is preserved or transformed in the generated response, we compare representations against the target emotion's contrastive direction at layer 11, the layer selected by the L2 probe. This direction is derived from the Set 1 emotion prototypes. For each emotion $e$, the prototype $\boldsymbol{\mu}_{e}^{(11)}$ is the mean of the 400 pooled layer-11 hidden states from Set 1 descriptions labeled with emotion $e$. The grand mean $\boldsymbol{\mu}^{(11)}$ is computed across all 4,000 Set 1 examples. The contrastive direction is the centered prototype vector:

\begin{equation}
\mathbf{d}_{e}^{(11)}
=
\boldsymbol{\mu}_{e}^{(11)}
-
\boldsymbol{\mu}^{(11)}.
\end{equation}

This vector captures what distinguishes emotion $e$ from the average emotion representation at the model's peak decoding layer. Cosine similarity was used to compare representations with this direction, making the measure invariant to vector magnitude.

For each generated response, three representations are extracted and compared:

\begin{enumerate}
    \item \textbf{Pooled input utterance.} The layer-11 pooled hidden state of the user's first utterance, computed with the same masked mean-pooling procedure used to construct the prototypes. This represents the affective profile of the input as a whole.
    
    \item \textbf{Pooled generated response.} The layer-11 pooled hidden state of the generated response text, computed after generation is complete. This represents the affective profile of the model's output as a whole.
    
    \item \textbf{Token-level response states.} The layer-11 hidden state of each generated token, taken in its full generation context at the moment the token is produced. These states trace how target-emotion alignment evolves token by token as the response unfolds.
\end{enumerate}

The pooled input and response representations are computed from raw text using the same pooling procedure applied to Set 1. The token-level states are computed in the full chat-template and generation context. Because these two kinds of representations differ in context and aggregation, their absolute cosine values are not directly comparable; we therefore analyze pooled and token-level trajectories separately. Target-emotion alignment is computed for each representation $\mathbf{h}$ as the cosine similarity to the target emotion's contrastive direction.


Emotion drift is defined as the difference between input alignment and response alignment:

\begin{equation}
\operatorname{Drift}_{e}
=
\cos\left(\mathbf{h}_{\mathrm{input}}, \mathbf{d}_{e}^{(11)}\right)
-
\cos\left(\mathbf{h}_{\mathrm{response}}, \mathbf{d}_{e}^{(11)}\right)
\end{equation}

Positive drift indicates that the generated response is less aligned with the target emotion than the input; negative drift indicates stronger alignment in the response. Drift thus quantifies the representational shift between the emotional content encoded in the input and the affectiveness produced in the output. Token-level trajectories are inspected to determine whether this shift arises gradually over the course of generation. 

For a relative category-level measure, each input and response representation was also compared with all 10 emotion directions. The argmax prediction was the emotion whose direction produced the highest cosine similarity. Target-emotion argmax accuracy was the proportion of examples for which this predicted emotion matched the dataset label (see Appendix~\ref{app:argmax}).

\section{Results}

\subsection{Emotion Representation Across Model Depth}
\label{sec:results-representation}

\begin{figure}[!htbp]
\centering
\includegraphics[width=\columnwidth, height=0.25\textheight, keepaspectratio]{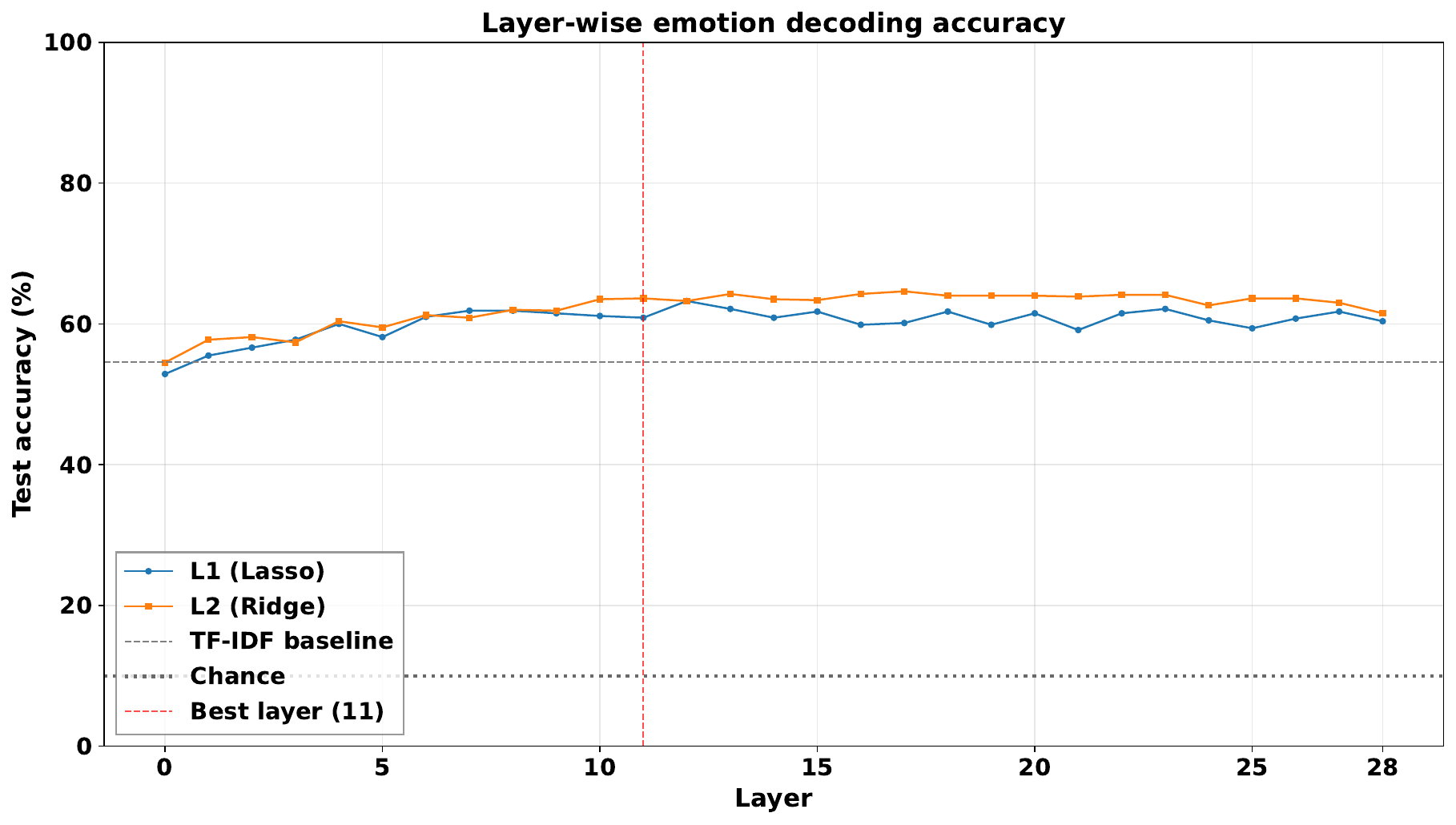}
\caption{Test emotion-classification accuracy of L1- and L2-regularized probes across representational levels. Lines denote mean stratified five-fold cross-validation accuracy. Horizontal lines indicate the TF-IDF and chance baselines. The vertical marker indicates layer 11, selected by L2 cross-validation.}
\label{fig:probing}
\end{figure}

Emotion-category information is linearly decodable from hidden representations at substantially above-chance levels. Cross-validation on the training partition selected layer 11 for the L2-regularized probe, which achieved a mean cross-validated accuracy of 65.6\% and a held-out accuracy of 63.6\% (macro-F1 = 0.635). The L1-regularized probe achieved 60.9\% held-out accuracy. Both probes substantially exceed the 10\% chance level and the 54.6\% TF-IDF lexical baseline (Figure~\ref{fig:probing}).

Accuracy increased from the embedding layer through the early transformer layers, peaked at layer 11, remained broadly comparable across layers 8--17, and declined toward the final layers. Emotion-category information is organized across a broad middle-layer region, not localized at any single depth.

The dense L2 probe achieved 63.6\% accuracy, compared with 60.9\% for the sparse L1 probe. Emotion-category distinctions are distributed across many partially redundant dimensions rather than confined to a small subset of features. Pairwise separability, prototype geometry, and probe-weight analyses are reported in the Supplementary Material (Appendix~\ref{app:probe-weights})

\subsection{Attention Allocation During Response Generation}
\label{sec:results-attention}

\begin{figure}[t]
\centering
\includegraphics[width=\columnwidth, height=0.25\textheight, keepaspectratio]{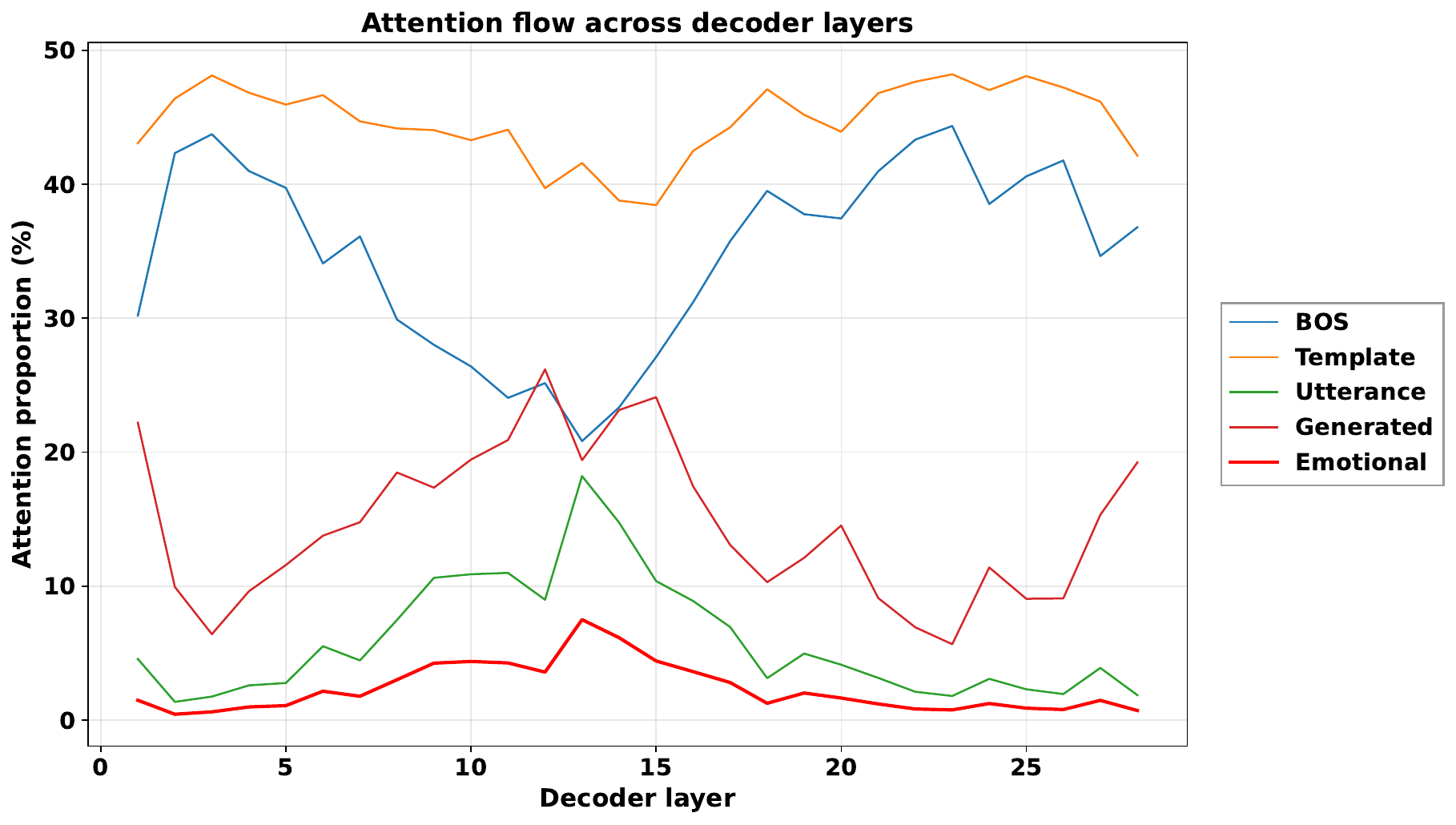}
\caption{Attention allocation across decoder layers and token-source groups. Lines denote sample-level means across 200 generated responses.}
\label{fig:attention-allocation}
\end{figure}

Averaged across responses, generation steps, layers, and attention heads, 44.7\% of attention was directed to template tokens, 34.8\% to the BOS token, 14.7\% to previously generated tokens, and 5.8\% to the user utterance. Within the utterance, emotion-related tokens received 1.89 times as much attention per token as non-emotional utterance tokens (Figure~\ref{fig:attention-allocation}). Attention to emotion-related tokens varied across model depth. Peak allocation occurred at layer 13 (7.5\%), with the next highest values at layers 14, 15, 10, and 11. This concentration across layers 10--15 descriptively overlaps with the broad middle-layer region of strongest emotion decodability.

\subsection{Input-Response Emotion Alignment}
\label{sec:results-alignment}

\begin{figure}[!htbp]
\centering
\includegraphics[width=\columnwidth, height=0.35\textheight, keepaspectratio]{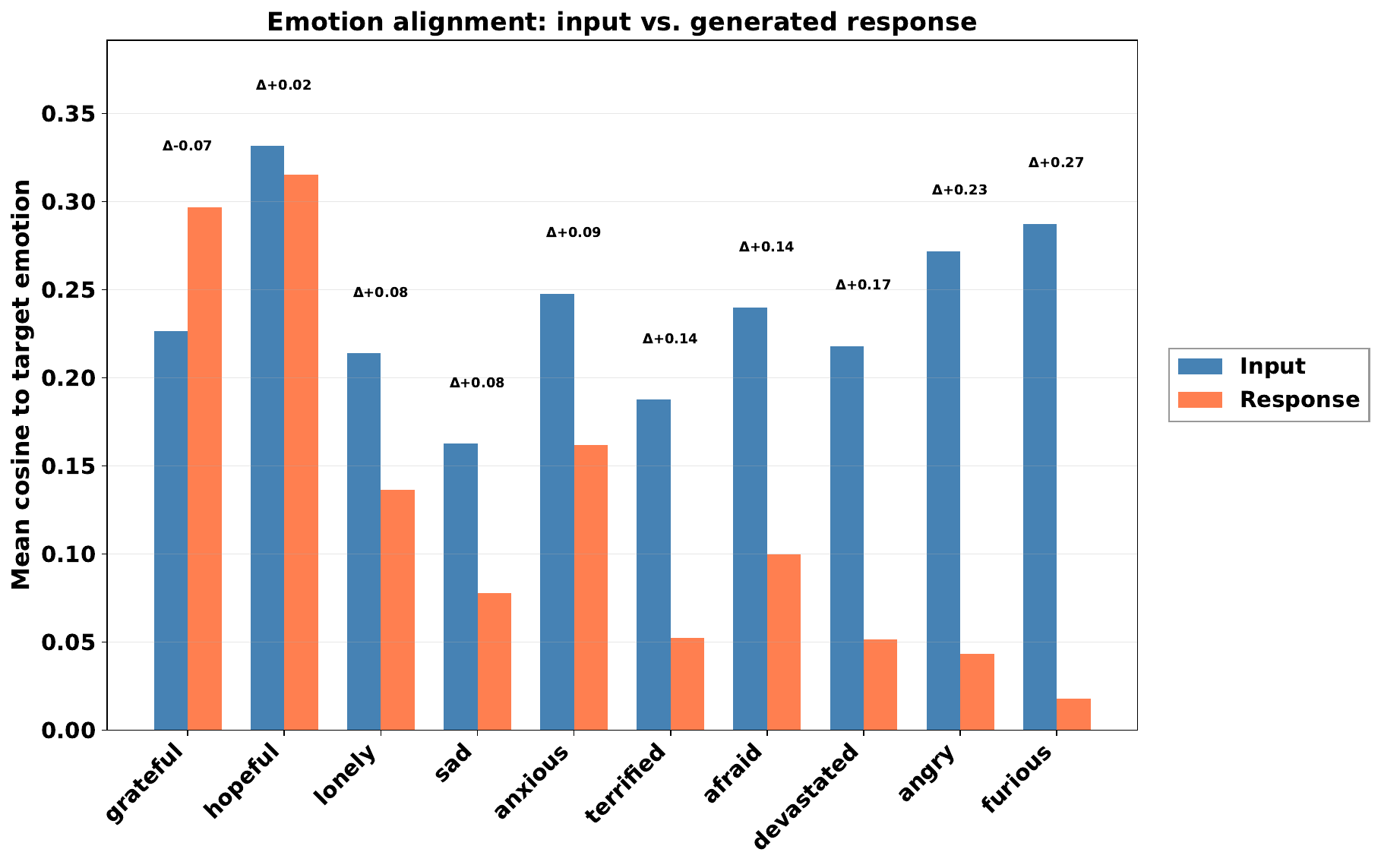}
\caption{Target-emotion cosine alignment of input utterances and generated responses. Bars denote emotion-specific means across 20 paired examples. Emotions are ordered by mean drift (input alignment minus response alignment).}
\label{fig:emotion-trajectory}
\end{figure}

The aggregate token-level trajectory showed lower target-emotion alignment at later response positions. Mean cosine similarity to the target-emotion direction was 0.239 for input utterances and 0.125 for generated responses, yielding a mean drift of 0.113. Positive drift was observed in 77\% of the 200 paired examples. Drift varied strongly by emotion category (Figure~\ref{fig:emotion-trajectory}). \emph{Grateful} exhibited negative drift ($-0.070$), meaning the response was more aligned with the target emotion than the input, and \emph{hopeful} showed near-zero drift ($0.017$). Positive drift increased across \emph{lonely} ($0.077$), \emph{sad} ($0.085$), \emph{anxious} ($0.086$), \emph{terrified} ($0.135$), \emph{afraid} ($0.140$), \emph{devastated} ($0.166$), \emph{angry} ($0.229$), and \emph{furious} ($0.269$).  The two positive-valence categories showed the strongest retention of target-emotion alignment, whereas the anger-related categories showed the largest attenuation. 

This pattern was confirmed by argmax classification: the target emotion was the highest-scoring direction emotion in 85\% of \emph{grateful} and 75\% of \emph{hopeful} response representations, compared with 0\% for \emph{devastated}, 10\% for \emph{furious}, and 20\% for both \emph{afraid} and \emph{terrified}. Token-level trajectories further showed that target-emotion alignment generally decreased toward the end of each generated response (see Figure~\ref{fig:response-token-trajectory} in Appendix~\ref{app:trajectories})

\section{Discussion}

The findings distinguish three components of emotion-related processing in Llama-3.2-3B-Instruct: emotion representation, attention to emotionally relevant input, and input--response alignment. Together, the results suggest that emotion-category information is internally accessible and selectively attended to during generation, but is not necessarily preserved in the model's response.

Emotion-category information was most accessible across a broad middle-layer region, with peak L2 probe performance at layer 11 and comparable performance across approximately layers 8--17. The sparse L1 probe retained 60.9\% held-out accuracy after setting 63.4\% of its coefficients to zero, compared with 63.6\% for the dense L2 probe. This suggests that emotion-category information is distributed across multiple hidden features rather than localized to a single dimension. Pairwise analyses further showed that semantically distinct emotions were more separable than closely related categories such as \emph{afraid} and \emph{terrified}, indicating that the internal structure reflects relationships among emotion categories. This middle-layer accessibility is consistent with recent probing results in Llama models \citep{palma2025llamas} and, at a functional level, parallels evidence that human emotions are represented through distributed rather than isolated neural patterns \citep{lindquist2012brain, saarimaki2018distributed}.

During generation, emotion-related utterance tokens received $1.89\times$ as much attention per token as non-emotional utterance tokens, with preferential attention concentrated across layers 10--15. The descriptive overlap between this region and the layers of strongest emotion decodability suggests a convergence between where emotion information is representationally accessible and where emotionally relevant content receives elevated attention during response construction. Thus, the model does not merely encode emotion-category information; it continues to preferentially allocate attention to emotion-relevant input as it generates a response. However, because attention reflects allocation rather than causal influence, intervention-based analyses are needed to establish whether these tokens or layers directly shape the response \citep{jain2019attention, wiegreffe2019attention}.

Despite this representational and attentional sensitivity, target-emotion alignment decreased from input to response. Mean cosine similarity fell from 0.239 to 0.125, with 77\% of examples showing positive drift. Importantly, this transformation was category dependent: alignment was retained or amplified for \emph{grateful} and largely preserved for \emph{hopeful}, whereas \emph{angry} and \emph{furious} showed the largest reductions. The argmax analysis showed the same tendency, with positive categories more frequently remaining the highest-scoring emotion direction, while several negative categories shifted toward other emotions. These results indicate that recognizing and attending to an emotional state does not imply that the same state will be preserved in the generated response.

This input-response shift may reflect two different processes. A supportive response may appropriately transform rather than reproduce the speaker's emotion; for example, responding to fear or anger with a calmer affective profile may be consistent with the listener's conversational role and with appraisal and regulation accounts of emotion \citep{lazarus1991emotion, riess2017}. Alternatively, the shift may reflect alignment-induced preferences for positive or neutral language, particularly under the medical-assistant prompt. 

The cognitive--affective empathy distinction provides a useful functional interpretation of these findings. Decodable emotion-category information provides a measurable analog of the representational component of cognitive empathy, while preferential attention indicates that emotion-relevant information remains salient during response construction. The subsequent input--response shift demonstrates that emotion representation and emotionally responsive generation are separable processes: the model can represent and attend to a speaker's emotion without simply reproducing it. This comparison is functional rather than experiential and does not imply human-like feeling or understanding. Instead, the framework decomposes emotion-aware generation into measurable stages: representation, attention, and response transformation, providing a basis for identifying where emotional information emerges, where it is prioritized, and how it changes as the model constructs a response.

\subsection{Implications}

Beyond characterizing emotion processing, these findings provide a mechanistic framework for evaluating emotion-aware generation in LLMs. By separating emotion representation, attention allocation, and response transformation, the framework can distinguish whether an emotionally inappropriate response arises from limited representation of the user's emotion, insufficient attention to emotion-relevant context, or a transformation that occurs during response generation. The overlap between emotion decodability and preferential attention in the middle layers further identifies candidate representations and model depths for future causal interventions. Rather than treating empathy as a property of the final response alone, these measures provide a basis for monitoring how emotional information is processed internally and for testing whether targeted interventions can improve the generation of emotionally appropriate responses without explicitly prescribing a particular emotion.

\subsection{Limitations and Future Directions}

The findings are correlational and scope-limited in several respects. First, linear probing demonstrates that emotion information is accessible from hidden states, but does not establish that these representations are causally used during generation \citep{hewitt2019designing, ravichander2021probing}. Similarly, attention indicates where the model allocates processing focus rather than causal influence \citep{jain2019attention, wiegreffe2019attention}. Future work could therefore use activation steering, ablation, or representation-engineering interventions to test whether the identified representations directly influence emotionally responsive generation \citep{zou2023representation}.

Second, EmpatheticDialogues labels describe the broader emotional situation, and the target emotion may not always be explicit in the isolated first utterance used for generation. Future studies could validate emotion labels at the utterance level or provide the full situational context to better align the model input with the target emotion.

Third, the generation analysis used 20 responses per emotion under a single medical-assistant prompt, and the emotion set was weighted toward negative and closely related categories. Future work could use larger samples, varied conversational roles, and more balanced or dimensional emotion sets to assess the generalizability of the observed emotional drift.

The current analyses can identify when and how emotional representations change between input and response, but cannot determine whether such transformations produce more empathetic responses. Human evaluation and comparison with human listener responses will therefore be important for distinguishing contextually appropriate emotional regulation from a general tendency toward positive or neutral responses.

Finally, the study focuses exclusively on Llama-3.2-3B-Instruct. Applying the same framework across model families, model scales, and alignment procedures would determine whether the observed middle-layer organization and input--response transformation reflect broader properties of LLM emotion processing or characteristics specific to this model.

\section{Conclusion}

This study examined how emotion-related information moves from input encoding to response generation in Llama-3.2-3B-Instruct. Layer-wise probing showed that emotion categories were most accessible across a broad middle-layer region, with the layer-11 L2 probe achieving 63.6\% held-out accuracy. During generation, emotion-related user tokens received $1.89\times$ as much attention per token as other utterance tokens, with the strongest allocation occurring in nearby middle layers. Despite this representational and attentional sensitivity, target-emotion alignment decreased from input to response in 77\% of examples, with the largest shifts for \emph{angry} and \emph{furious}. Together, these findings distinguish three components of emotion processing in LLMs: representing a user's emotional state, attending to emotionally relevant content, and transforming that information during response generation. This framework provides a mechanistic basis for tracking emotion-related information across model depth and identifying candidate representations for future causal interventions aimed at improving emotionally appropriate generation.

\section{AI Use Statement}

Generative AI tools were used to assist with language editing, clarity, and polishing of the manuscript. The authors developed the research questions, methodology, analyses, interpretation of results, and scientific conclusions. All AI-assisted text was reviewed and revised by the authors, who take full responsibility for the final content.

\clearpage

\bibliographystyle{unsrt}
\bibliography{reference}

@book{lazarus1991emotion,
  author    = {Lazarus, Richard S.},
  title     = {Emotion and Adaptation},
  publisher = {Oxford University Press},
  address   = {New York, NY},
  year      = {1991}
}

@article{lindquist2012brain,
  author  = {Lindquist, Kristen A. and Wager, Tor D. and Kober, Hedy and
             Bliss-Moreau, Eliza and Barrett, Lisa Feldman},
  title   = {The Brain Basis of Emotion: {A} Meta-Analytic Review},
  journal = {Behavioral and Brain Sciences},
  volume  = {35},
  number  = {3},
  pages   = {121--143},
  year    = {2012},
  doi     = {10.1017/S0140525X11000446}
}

@article{riess2017,
  author  = {Riess, Helen},
  title   = {The Science of Empathy},
  journal = {Journal of Patient Experience},
  volume  = {4},
  number  = {2},
  pages   = {74--77},
  year    = {2017},
  doi     = {10.1177/2374373517699267}
}

@article{fan2011,
  author  = {Fan, Yan and Duncan, Niall W. and de Greck, Moritz and Northoff, Georg},
  title   = {Is There a Core Neural Network in Empathy? {An} {fMRI} Based
             Quantitative Meta-Analysis},
  journal = {Neuroscience \& Biobehavioral Reviews},
  volume  = {35},
  number  = {3},
  pages   = {903--911},
  year    = {2011},
  doi     = {10.1016/j.neubiorev.2010.10.009}
}

@article{saarimaki2018distributed,
  author  = {Saarim{\"a}ki, Heini and Ejtehadian, Lara Farzaneh and
             Glerean, Enrico and J{\"a}{\"a}skel{\"a}inen, Iiro P. and
             Vuilleumier, Patrik and Sams, Mikko and Nummenmaa, Lauri},
  title   = {Distributed Affective Space Represents Multiple Emotion
             Categories across the Human Brain},
  journal = {Social Cognitive and Affective Neuroscience},
  volume  = {13},
  number  = {5},
  pages   = {471--482},
  year    = {2018},
  doi     = {10.1093/scan/nsy018}
}

@article{baucom2012decoding,
  author  = {Baucom, Laura B. and Wedell, Douglas H. and Wang, Jing and
             Blitzer, David and Shinkareva, Svetlana V.},
  title   = {Decoding the Neural Representation of Affective States},
  journal = {NeuroImage},
  volume  = {59},
  number  = {1},
  pages   = {718--727},
  year    = {2012},
  doi     = {10.1016/j.neuroimage.2011.07.037}
}

@article{fisher1936use,
  author  = {Fisher, Ronald A.},
  title   = {The Use of Multiple Measurements in Taxonomic Problems},
  journal = {Annals of Eugenics},
  volume  = {7},
  number  = {2},
  pages   = {179--188},
  year    = {1936},
  doi     = {10.1111/j.1469-1809.1936.tb02137.x}
}

@article{fitzpatrick2017,
  author  = {Fitzpatrick, Kathleen Kara and Darcy, Alison and Vierhile, Molly},
  title   = {Delivering Cognitive Behavior Therapy to Young Adults with Symptoms
             of Depression and Anxiety Using a Fully Automated Conversational
             Agent ({W}oebot): {A} Randomized Controlled Trial},
  journal = {JMIR Mental Health},
  volume  = {4},
  number  = {2},
  pages   = {e19},
  year    = {2017},
  doi     = {10.2196/mental.7785}
}

@inproceedings{sharma2021,
  author    = {Sharma, Ashish and Lin, Inna W. and Miner, Adam S. and
               Atkins, David C. and Althoff, Tim},
  title     = {Towards Facilitating Empathic Conversations in Online Mental
               Health Support: {A} Reinforcement Learning Approach},
  booktitle = {Proceedings of the Web Conference 2021 (WWW '21)},
  pages     = {1817--1829},
  year      = {2021},
  publisher = {ACM},
  doi       = {10.1145/3442381.3450097}
}

@inproceedings{rashkin2019towards,
  author    = {Rashkin, Hannah and Smith, Eric Michael and Li, Margaret and
               Boureau, Y-Lan},
  title     = {Towards Empathetic Open-domain Conversation Models:
               {A} New Benchmark and Dataset},
  booktitle = {Proceedings of the 57th Annual Meeting of the Association for
               Computational Linguistics},
  pages     = {5370--5381},
  year      = {2019},
  address   = {Florence, Italy},
  publisher = {Association for Computational Linguistics},
  doi       = {10.18653/v1/P19-1534}
}

@inproceedings{lee2024large,
  author    = {Lee, Yoon Kyung and Suh, Jina and Zhan, Hongli and
               Li, Junyi Jessy and Ong, Desmond C.},
  title     = {Large Language Models Produce Responses Perceived to be Empathic},
  booktitle = {Proceedings of the 12th International Conference on Affective
               Computing and Intelligent Interaction (ACII)},
  pages     = {63--71},
  year      = {2024},
  address   = {Glasgow, United Kingdom},
  publisher = {IEEE},
  url       = {https://arxiv.org/abs/2403.18148}
}

@inproceedings{maheswaran2025,
  author    = {Maheswaran, Aishwarya and Chua, Caslon and
               Desarkar, Maunendra Sankar},
  title     = {Probing the Inherent Ability of Large Language Models for
               Generating Empathetic Responses},
  booktitle = {Proceedings of the IEEE Swiss Conference on Data Science (SDS)},
  pages     = {32--39},
  year      = {2025},
  address   = {Z{\"u}rich, Switzerland},
  publisher = {IEEE}
}

@article{grattafiori2024llama,
  author  = {Grattafiori, Aaron and Dubey, Abhimanyu and Jauhri, Abhinav and
             Pandey, Abhinav and Kadian, Abhishek and Al-Dahle, Ahmad and
             Kelsey, Aiesha and others},
  title   = {The {Llama 3} Herd of Models},
  journal = {arXiv preprint arXiv:2407.21783},
  year    = {2024},
  url     = {https://arxiv.org/abs/2407.21783}
}

@misc{meta2024llama32,
  author       = {{Meta AI}},
  title        = {{Llama 3.2} Model Card},
  year         = {2024},
  howpublished = {GitHub},
  url          = {https://github.com/meta-llama/llama-models/blob/main/models/llama3_2/MODEL_CARD.md},
  note         = {Released October 2024}
}

@inproceedings{alain2017understanding,
  author    = {Alain, Guillaume and Bengio, Yoshua},
  title     = {Understanding Intermediate Layers Using Linear Classifier Probes},
  booktitle = {International Conference on Learning Representations (Workshop Track)},
  year      = {2017},
  url       = {https://arxiv.org/abs/1610.01644}
}

@article{belinkov2022probing,
  author  = {Belinkov, Yonatan},
  title   = {Probing Classifiers: Promises, Shortcomings, and Advances},
  journal = {Computational Linguistics},
  volume  = {48},
  number  = {1},
  pages   = {207--219},
  year    = {2022},
  doi     = {10.1162/COLI_a_00422}
}

@inproceedings{hewitt2019designing,
  author    = {Hewitt, John and Liang, Percy},
  title     = {Designing and Interpreting Probes with Control Tasks},
  booktitle = {Proceedings of the 2019 Conference on Empirical Methods in
               Natural Language Processing and the 9th International Joint
               Conference on Natural Language Processing (EMNLP-IJCNLP)},
  pages     = {2733--2743},
  year      = {2019},
  address   = {Hong Kong, China},
  publisher = {Association for Computational Linguistics},
  doi       = {10.18653/v1/D19-1275}
}

@inproceedings{ravichander2021probing,
  author    = {Ravichander, Abhilasha and Belinkov, Yonatan and Hovy, Eduard},
  title     = {Probing the Probing Paradigm: Does Probing Accuracy Entail
               Task Relevance?},
  booktitle = {Proceedings of the 16th Conference of the European Chapter of
               the Association for Computational Linguistics: Main Volume},
  pages     = {3363--3377},
  year      = {2021},
  address   = {Online},
  publisher = {Association for Computational Linguistics},
  doi       = {10.18653/v1/2021.eacl-main.295}
}

@inproceedings{palma2025llamas,
  author    = {Di Palma, Dario and De Bellis, Alessandro and Servedio, Giovanni
               and Anelli, Vito Walter and Narducci, Fedelucio and
               Di Noia, Tommaso},
  title     = {{LL}a{MA}s Have Feelings Too: Unveiling Sentiment and Emotion
               Representations in {LL}a{MA} Models Through Probing},
  booktitle = {Proceedings of the 63rd Annual Meeting of the Association for
               Computational Linguistics (Volume 1: Long Papers)},
  pages     = {6124--6142},
  year      = {2025},
  address   = {Vienna, Austria},
  publisher = {Association for Computational Linguistics},
  doi       = {10.18653/v1/2025.acl-long.306}
}

@inproceedings{reimers2019sentence,
  author    = {Reimers, Nils and Gurevych, Iryna},
  title     = {Sentence-{BERT}: Sentence Embeddings using {S}iamese
               {BERT}-Networks},
  booktitle = {Proceedings of the 2019 Conference on Empirical Methods in
               Natural Language Processing and the 9th International Joint
               Conference on Natural Language Processing (EMNLP-IJCNLP)},
  pages     = {3982--3992},
  year      = {2019},
  address   = {Hong Kong, China},
  publisher = {Association for Computational Linguistics},
  doi       = {10.18653/v1/D19-1410}
}

@article{zou2023representation,
  author  = {Zou, Andy and Phan, Long and Chen, Sarah and Campbell, James and
             Guo, Phillip and Ren, Richard and Pan, Alexander and Yin, Xuwang
             and Mazeika, Mantas and Dombrowski, Ann-Kathrin and Goel, Shashwat
             and Li, Nathaniel and Byun, Michael J. and Wang, Zifan and
             Mallen, Alex and Basart, Steven and Koyejo, Sanmi and Song, Dawn
             and Fredrikson, Matt and Kolter, J. Zico and Hendrycks, Dan},
  title   = {Representation Engineering: {A} Top-Down Approach to {AI}
             Transparency},
  journal = {arXiv preprint arXiv:2310.01405},
  year    = {2023},
  url     = {https://arxiv.org/abs/2310.01405}
}

@inproceedings{jain2019attention,
  author    = {Jain, Sarthak and Wallace, Byron C.},
  title     = {Attention is not Explanation},
  booktitle = {Proceedings of the 2019 Conference of the North American Chapter
               of the Association for Computational Linguistics: Human Language
               Technologies, Volume 1 (Long and Short Papers)},
  pages     = {3543--3556},
  year      = {2019},
  address   = {Minneapolis, Minnesota},
  publisher = {Association for Computational Linguistics},
  doi       = {10.18653/v1/N19-1357}
}

@inproceedings{wiegreffe2019attention,
  author    = {Wiegreffe, Sarah and Pinter, Yuval},
  title     = {Attention is not not Explanation},
  booktitle = {Proceedings of the 2019 Conference on Empirical Methods in
               Natural Language Processing and the 9th International Joint
               Conference on Natural Language Processing (EMNLP-IJCNLP)},
  pages     = {11--20},
  year      = {2019},
  address   = {Hong Kong, China},
  publisher = {Association for Computational Linguistics},
  doi       = {10.18653/v1/D19-1002}
}

\clearpage

\appendix
\section{Supplementary Material}

\begin{table*}[h]
\centering
\small
\begin{tabular}{p{0.12\textwidth} p{0.78\textwidth}}
\toprule
Emotion & First utterance \\
\midrule

Afraid &
I have a solo trip coming up. I'm going to Colorado to climb Mt. Elbert Solo. It's the tallest mountain in the United States.\\

\addlinespace

Angry &
On my way to work, a goose stole the toupee I needed to impress the new secretary, Christy. \\

\addlinespace

Anxious &
I have an important exam tomorrow and feel that something bad will happen. \\

\addlinespace

Devastated &
I had the worst night last night. I found out my uncle hit my cat with his car. It happened two weeks ago, and he never told me! \\

\addlinespace

Lonely &
I'm an over-the-road trucker. Man, it gets boring in the evenings. \\

\addlinespace

Sad &
Hello, yesterday was my birthday and my wife completely forgot! \\

\addlinespace

Terrified &
I moved into this new house and was woken up with this horrible screeching noise in the middle of the night. \\

\addlinespace

Furious &
I spent 5 hours baking a cake yesterday. \\

\addlinespace

Grateful &
My car broke down on my way to work yesterday. \\

\addlinespace

Hopeful &
I have been talking to this girl that I like for awhile now. I think that I am finally going to ask her on a date \\

\bottomrule
\end{tabular}
\caption{Examples of emotion-labeled first utterances from Set 2. First utterances were used as inputs to the model during response generation. Set 2 did not overlap with Set 1. Text is reproduced from the dataset, with capitalization added only where necessary for consistent presentation.}
\label{tab:set2-utterance-examples}
\end{table*}

\subsection{Zero-Shot Response-Generation Prompt}
\label{app:generation-prompt}

Each selected Set 2 utterance was inserted as the user message in the model's chat template. The following system prompt was used without demonstrations:

\begin{quote}
\ttfamily
You are a medical assistant. Given the user's emotional situation, do two things:

1. Write a response (1--2 sentences).

2. Rate how emotional is the user's situation on a scale of 1 (mild) to 5 (highly emotional).

Format your output exactly as:

Response: <your response>

Intensity: <1--5>
\end{quote}

Responses were generated autoregressively using nucleus sampling with a temperature of 0.7 and top-$p$ of 0.9, with a maximum of 50 new tokens and early termination upon generation of the end-of-sequence token. The system prompt and user utterance were formatted using the tokenizer's model-specific chat template with the generation prompt enabled.

\subsection{Linear-Probe Implementation Details}
\label{app:probing}
Two regularization schemes were evaluated:

\begin{itemize}
    \item \textbf{L1 regularization:} $\lambda \in
    \{10^{-7}, 10^{-6.5}, \ldots, 10^{-2}\}$.
    L1 probes were optimized using proximal gradient descent, producing exact zero-valued coefficients.
    
    \item \textbf{L2 regularization:} $\lambda \in
    \{10^{-4},10^{-3},\ldots,10^{4}\}$.
    L2 probes were optimized using full-batch limited-memory BFGS.
\end{itemize}

The regularization strength was selected independently at each representational level using stratified 5-fold cross-validation on the training partition. The best representational level was selected using cross-validated accuracy. The held-out test set was evaluated only after model and layer selection.

\subsection{Lexical and Control Baselines}
\label{app:baselines}

Four baselines contextualize probe performance. The \textbf{TF-IDF baseline} used multinomial logistic regression on lexical features extracted from Set 1 descriptions. The \textbf{embedding-level baseline} applied the L2 probing procedure to the initial token-embedding representations. The \textbf{nearest-centroid baseline} assigned each test example to the nearest emotion centroid estimated from the standardized training representations. The \textbf{shuffled-label control} repeated probe training with randomly permuted training labels. All baselines used the same train-test partition, and applicable hyperparameters were selected by cross-validation (Table~\ref{tab:probe-baselines}).

\begin{table*}[h]
\centering

\begin{minipage}[t]{0.47\textwidth}
\vspace{0pt}
\centering
\small
\begin{tabular}{lr}
\toprule
Method & Accuracy (\%) \\
\midrule
Chance & 10.0 \\
Shuffled-label control & 10.6 \\
Nearest centroid, selected layer & 52.1 \\
Embedding-level L2 probe & 54.5 \\
TF--IDF logistic regression & 54.6 \\
Layer-11 L1 probe & 60.9 \\
Layer-11 L2 probe & \textbf{63.6} \\
\bottomrule
\end{tabular}
\captionof{table}{Held-out emotion-classification accuracy of the primary probes and comparison baselines.}
\label{tab:probe-baselines}
\end{minipage}
\hfill
\begin{minipage}[t]{0.47\textwidth}
\vspace{0pt}
\centering
\small
\begin{tabular}{rrr}
\toprule
Examples & CV & Accuracy \\
Per Emotion & (\%) & (\%) \\
\midrule 
80  & $61.0 \pm 2.8$ & 58.7 \\
160 & $61.1 \pm 1.7$ & 61.6 \\
240 & $63.7 \pm 1.9$ & 63.2 \\
320 & $65.6 \pm 2.5$ & 63.6 \\
\bottomrule
\end{tabular}
\captionof{table}{L2 probe performance by training-set size.}
\label{tab:learning-curve}
\end{minipage}

\end{table*}

\subsection{Learning-Curve Analysis}
\label{app:learning-curve}

L2 probes were trained using 80, 160, 240, and 320 examples per emotion while retaining a fixed held-out test partition. Held-out accuracy improved by 4.9 percentage points from 80 to 320 examples per emotion; the final increment (240 to 320) yielded only 0.4 points, indicating diminishing returns from additional labeled examples (Table~\ref{tab:learning-curve}).

\subsection{Emotion-Specific Performance and Pairwise Separability}
\label{app:per-emotion-recall}

Per-emotion recall was calculated for the held-out predictions of the layer-11 L2 probe. Binary L2 probes were also trained for all 45 emotion pairs, evaluated by cross-validation, and compared with TF-IDF classifiers trained on the same pairs. The weakest pairwise distinction was \emph{afraid}--\emph{terrified}, for which the hidden-state probe did not outperform the lexical baseline. More semantically distinct pairs were separated with substantially higher accuracy (Table~\ref{tab:pairwise-separability}).

\begin{table}[h]
\centering
\begin{tabular}{lrrr}
\toprule
Emotion pair & Hidden state & TF--IDF & Gain \\
\midrule
Afraid--terrified & 60.3 & 61.4 & $-1.1$ \\
Angry--furious & 60.9 & 57.2 & $+3.7$ \\
Devastated--sad & 75.5 & 66.7 & $+8.7$ \\
Lonely--furious & 98.8 & 92.7 & $+6.1$ \\
Angry--lonely & 98.3 & 91.7 & $+6.6$ \\
Furious--grateful & 98.0 & 87.5 & $+10.5$ \\
Sad--hopeful & 97.5 & 88.1 & $+9.4$ \\
\bottomrule
\end{tabular}
\caption{Selected pairwise emotion-classification results. ``Gain'' is the difference between hidden-state and TF-IDF cross-validated accuracy.}
\label{tab:pairwise-separability}
\end{table}

\subsection{Prototype Geometry}
\label{app:fisher}

For each emotion $e$ and representational level $\ell$, the emotion prototype was defined as

\begin{equation}
\boldsymbol{\mu}_{e}^{(\ell)}
=
\frac{1}{N_e}
\sum_{i:y_i=e}
\mathbf{h}_{i}^{(\ell)},
\end{equation}

where $\mathbf{h}_{i}^{(\ell)}$ is the pooled representation of example $i$ at level $\ell$. The grand mean was defined across all Set 1 examples:

\begin{equation}
\boldsymbol{\mu}^{(\ell)}
=
\frac{1}{N}
\sum_{i=1}^{N}
\mathbf{h}_{i}^{(\ell)}.
\end{equation}

The target-emotion contrastive direction was

\begin{equation}
\mathbf{d}_{e}^{(\ell)}
=
\boldsymbol{\mu}_{e}^{(\ell)}
-
\boldsymbol{\mu}^{(\ell)}.
\end{equation}

Raw prototype cosine similarities exceeded 0.99 across layers, reflecting a large component shared by all situation representations; raw cosine similarity was therefore not used as the primary measure of emotion-category separation. Representational separability was summarized using Fisher's discriminant criterion \citep{fisher1936use}:

\begin{equation}
F^{(\ell)}
=
\frac{\operatorname{tr}\left(S_B^{(\ell)}\right)}
     {\operatorname{tr}\left(S_W^{(\ell)}\right)},
\end{equation}

where $S_B^{(\ell)}$ and $S_W^{(\ell)}$ denote between-emotion and within-emotion scatter at level $\ell$. Fisher's ratio increased from 0.0278 at the embedding level to a maximum of 0.0862 at layer 13, indicating the strongest geometric separation near---but not at---the layer selected by the linear probe.

\begin{figure}[!htbp]
\centering
\includegraphics[width=\columnwidth, height = 0.32\textheight, keepaspectratio]{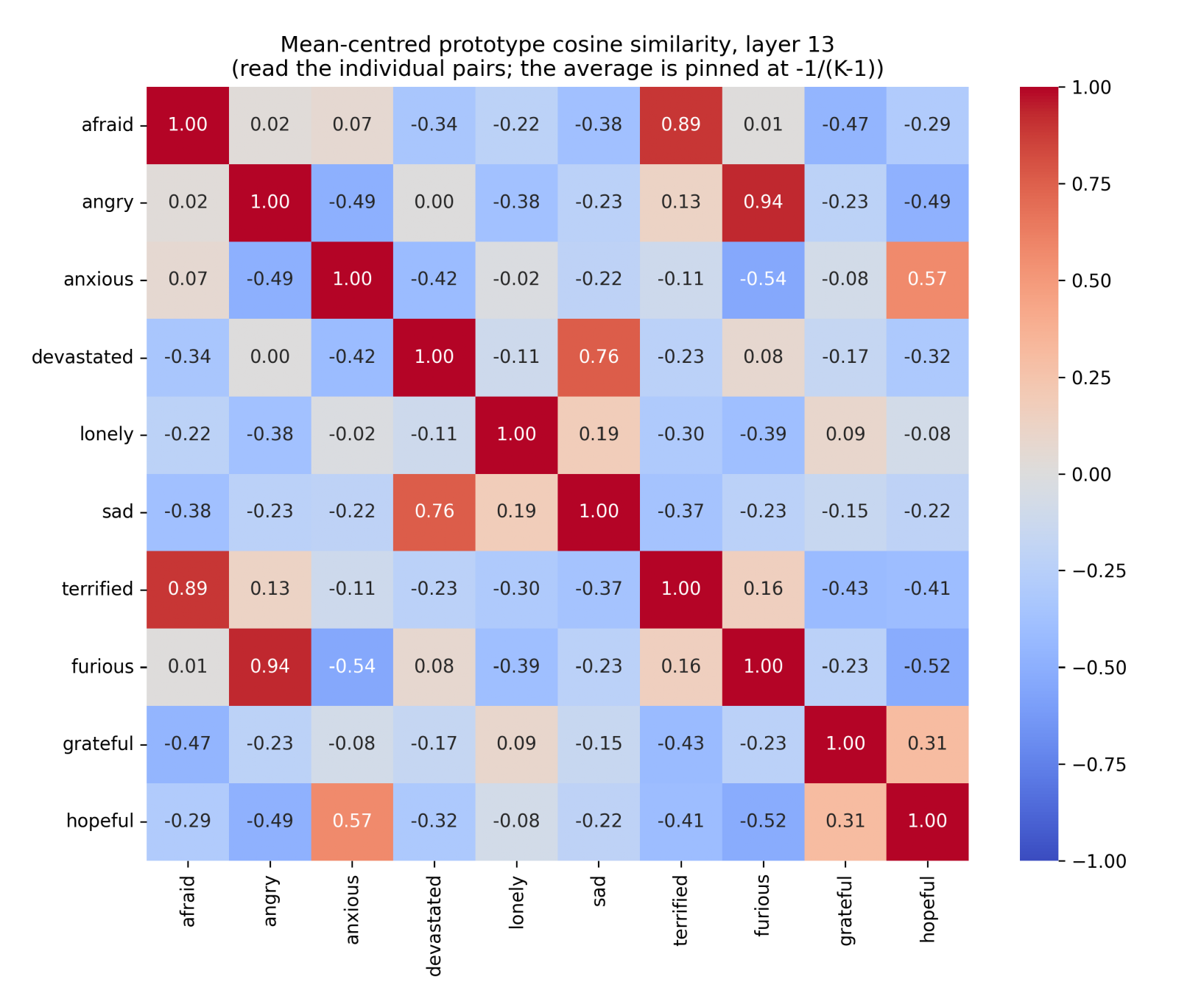}
\caption{Prototype geometry across representational levels. Raw emotion prototypes share a large common component; Fisher's ratio shows the strongest relative category separation at layer 13.}
\label{fig:prototype-geometry}
\end{figure}

\subsection{Probe-Weight Analysis}
\label{app:probe-weights}

\begin{figure*}[h]
\centering
\begin{minipage}[t]{0.43\textwidth}
\vspace{0pt}
\centering
\small
\begin{tabular}{lr}
\toprule
Emotion pair & Cosine similarity \\
\midrule
Afraid--terrified & 0.446 \\
Angry--furious & 0.411 \\
Sad--devastated & 0.213 \\
Grateful--hopeful & 0.087 \\
Anxious--afraid & 0.079 \\
\bottomrule
\end{tabular}
\captionof{table}{Selected layer-11 L2 probe-direction similarities.}
\label{tab:probe-direction-similarity}
\end{minipage}
\hfill
\begin{minipage}[t]{0.52\textwidth}
\vspace{0pt}
\centering
\includegraphics[width=\linewidth]{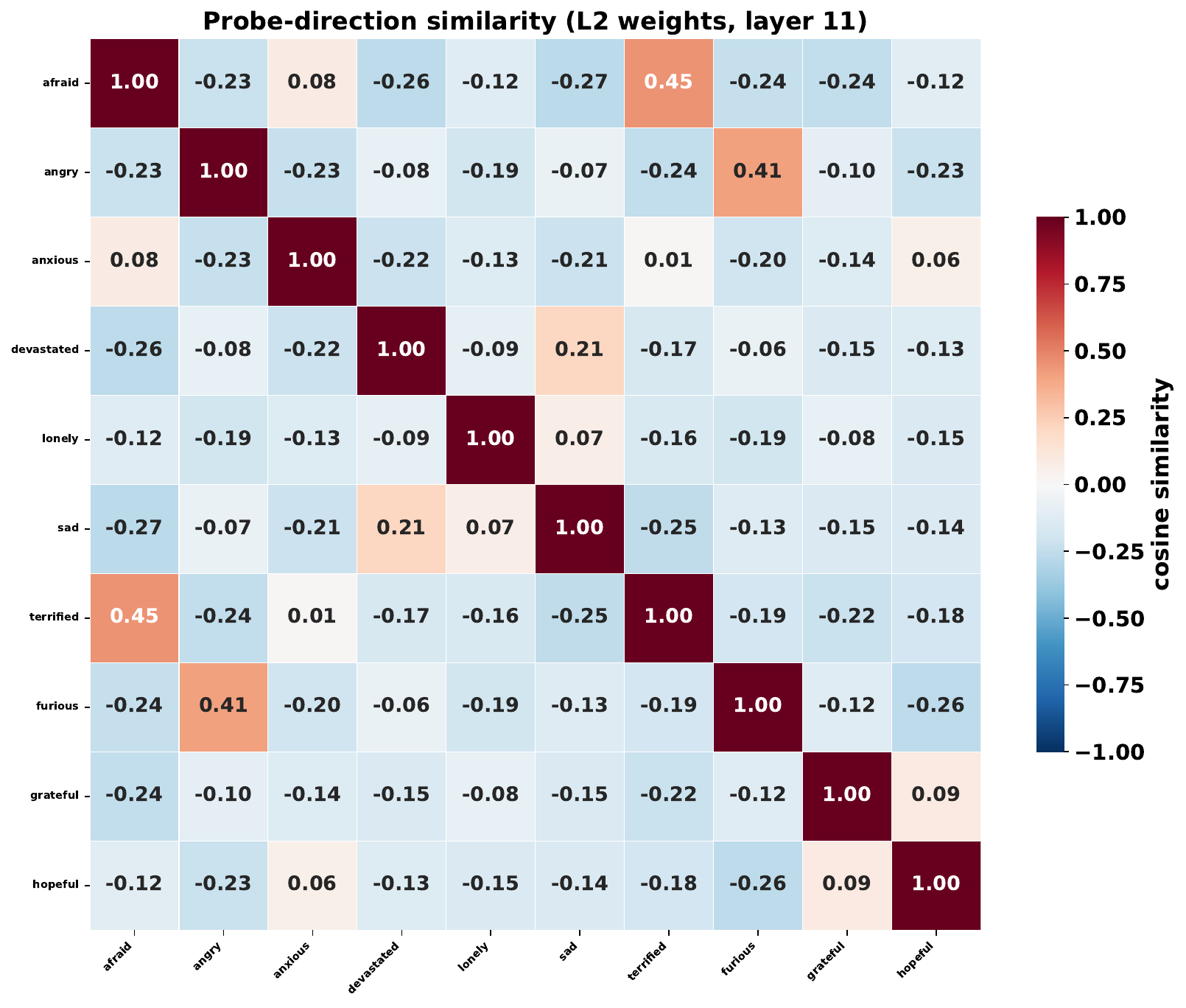}
\captionof{figure}{Similarity between emotion-specific L2 probe directions.}
\label{fig:probe-direction-similarity}
\end{minipage}
\end{figure*}

At layer 11, the sparse L1 probe set 63.4\% of its coefficients to exactly zero while retaining 60.9\% held-out accuracy. The number of nonzero coefficients per emotion ranged from 858 (lonely) to 1,318 (afraid) out of 3,072 possible dimensions. No single dimension appeared among the 20 highest-weighted dimensions for all 10 emotions. Cosine similarity between emotion-specific L2 weight vectors was highest for closely related pairs: \emph{afraid}--\emph{terrified} (0.446) and \emph{angry}--\emph{furious} (0.411), consistent with their shared affective structure (Table~\ref{tab:probe-direction-similarity}, Figure~\ref{fig:probe-direction-similarity}).

\subsection{Response-Emotion Trajectories}
\label{app:trajectories}

\begin{figure}[!htbp]
\centering
\includegraphics[width=\columnwidth, height = 0.25\textheight, keepaspectratio]{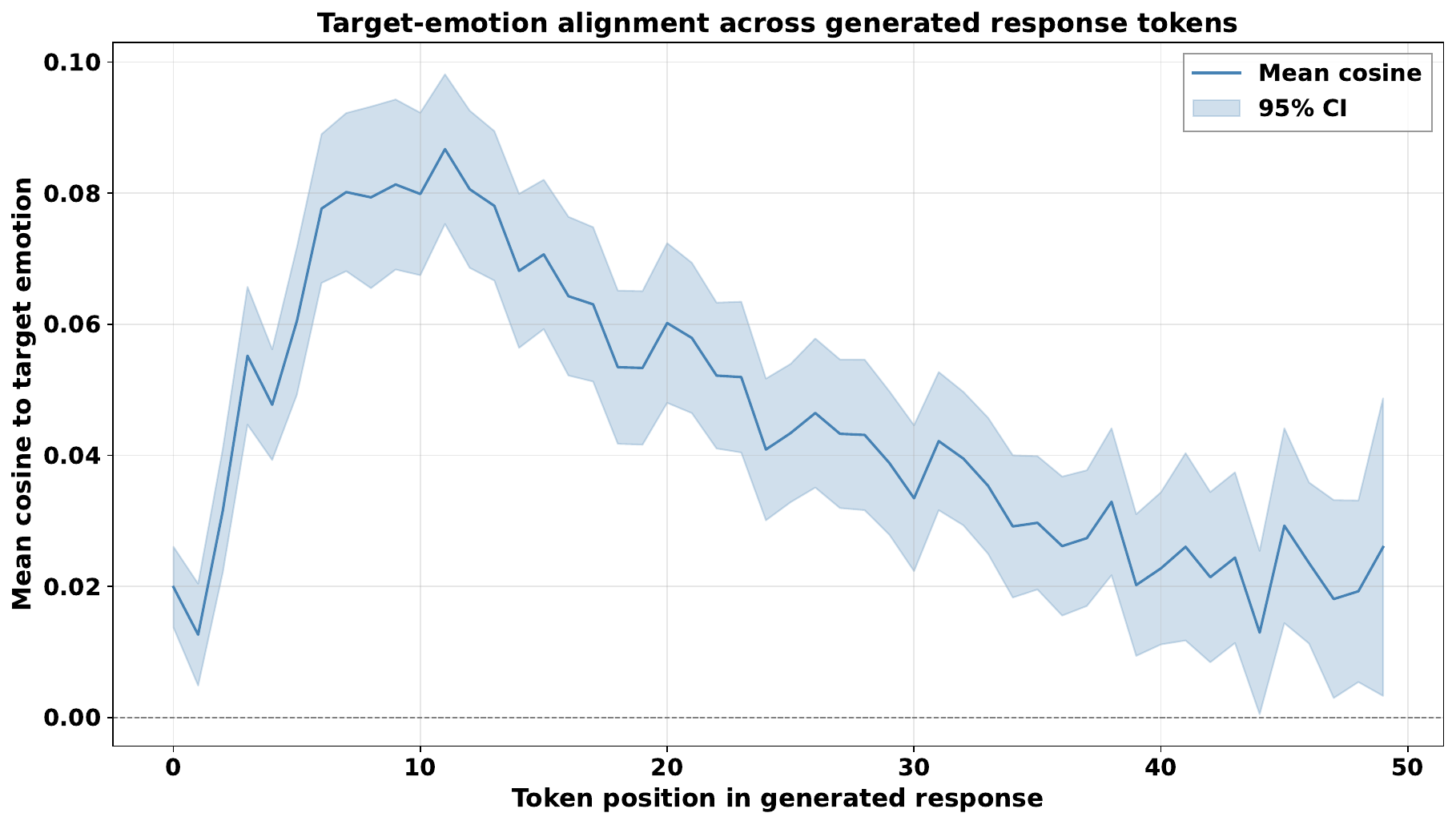}
\caption{Target-emotion alignment across generated-token positions. Token-level layer-11 hidden states were compared with the corresponding target-emotion direction in the full generation context. Lines show mean alignment across responses, and shaded regions show 95\% bootstrap confidence intervals.}
\label{fig:response-token-trajectory}
\end{figure}

Response-emotion trajectories were calculated at layer 11 using the target emotion's contrastive direction. Three representations were analyzed for each response: the pooled input utterance, the pooled generated response, and the contextual hidden state of each generated token. Pooled and token-level cosine values were analyzed separately and are not directly comparable in scale.

Token-level alignment was summarized across normalized response positions to account for differences in response length. As shown in Figure~\ref{fig:response-token-trajectory}, mean target-emotion alignment [insert the exact observed pattern and supporting values]. These contextual token-level values were analyzed separately from the pooled input and response representations because they differ in both context and aggregation.

\subsection{Target-Emotion Argmax Classification}
\label{app:argmax}

To complement the continuous cosine-alignment analysis, each pooled input and response representation was compared with all 10 layer-11 emotion directions. For a representation $\mathbf{h}$, the predicted emotion was defined as the direction with the highest cosine similarity:

\begin{equation}
\hat{e}(\mathbf{h})
=
\underset{e \in \mathcal{E}}{\operatorname{argmax}}
\;
\cos\left(\mathbf{h},\mathbf{d}_{e}^{(11)}\right),
\end{equation}

where $\mathcal{E}$ contains the 10 emotion categories and $\mathbf{d}_{e}^{(11)}$ is the layer-11 contrastive direction for emotion $e$. Target-emotion argmax accuracy was the proportion of examples for which $\hat{e}(\mathbf{h})$ matched the dataset label. This measure evaluates the target emotion relative to all other emotion directions, whereas the drift analysis measures the change in continuous alignment with the target direction alone.

Across all 200 examples, the target emotion was the highest-scoring direction for 50.5\% of input representations and 36.5\% of response representations. The mean rank of the target emotion among the 10 directions changed from 2.33 for inputs to 3.15 for responses, with lower values indicating stronger relative alignment. At the response level, argmax accuracy was highest for \emph{grateful} (85\%) and \emph{hopeful} (75\%). Lower values were observed for \emph{devastated} (0\%), \emph{furious} (10\%), and both \emph{afraid} and \emph{terrified} (20\%). These category-level results are consistent with the cosine-drift analysis: positive emotions retained stronger relative alignment, while several negative emotions shifted toward other directions in the response space.

\begin{figure}[!htbp]
\centering
\includegraphics[width=\columnwidth, height = 0.25\textheight, keepaspectratio]{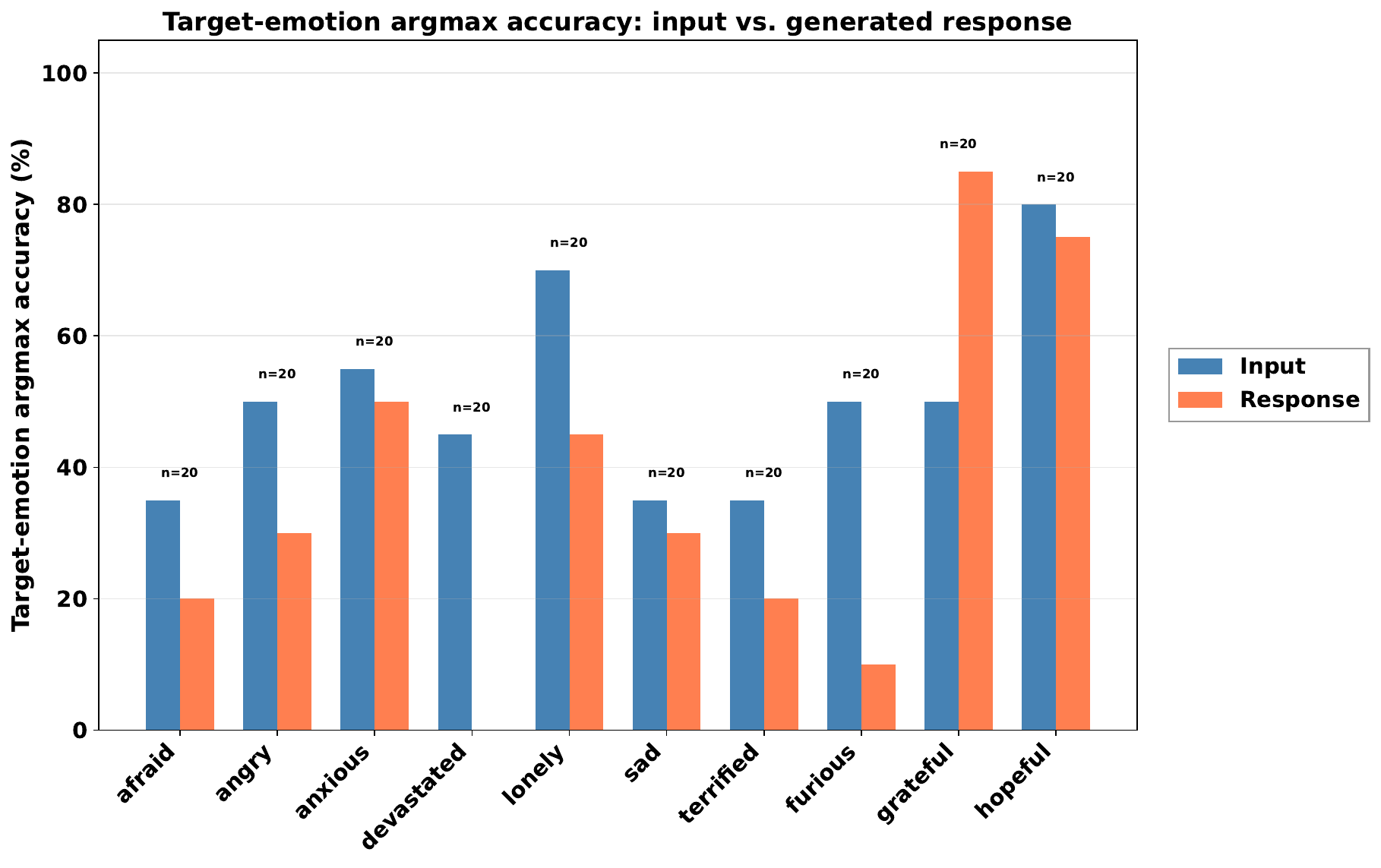}
\caption{Target-emotion argmax accuracy for pooled input and generated-response representations across the 10 emotion categories ($n=20$ per emotion). For each representation, the predicted emotion was the direction with the highest cosine similarity among the 10 layer-11 contrastive directions. Bars denote the proportion of examples for which the predicted emotion matched the dataset label.}
\label{fig:argmax-classification}
\end{figure}

\subsection{Reproducibility}

All dataset splitting and sampling procedures used a fixed random seed of 42. Probe hyperparameters were selected using only the training partition, and the held-out test partition was evaluated after model and layer selection.

\end{document}